# A game-theoretic framework for classifier ensembles using weighted majority voting with local accuracy estimates

Harris V. Georgiou[1]  &  Michael E. Mavroforakis[2]

**Abstract**

In this paper, a novel approach for the optimal combination of binary classifiers is proposed. The classifier combination problem is approached from a Game Theory perspective. The proposed framework of adapted weighted majority rules (WMR) is tested against common rank-based, Bayesian and simple majority models, as well as two soft-output averaging rules. Experiments with ensembles of Support Vector Machines (SVM), Ordinary Binary Tree Classifiers (OBTC) and weighted *k*-nearest-neighbor (w/*k*-NN) models on benchmark datasets indicate that this new *adaptive* WMR model, employing *local accuracy estimators* and the *analytically* computed optimal weights outperform all the other simple combination rules.

## 1. Introduction

Classifier combination is one of the most active areas of research in the discipline of Pattern Recognition. The challenging problem of designing optimal aggregation schemes for multi-classifier systems has been addressed by a wide range of methodologies and approaches during the last decade [1]. However, few of them introduce a framework of analytical solutions. Instead, most of them employ either heuristics or iterative optimization procedures.

In this paper, a novel viewpoint is proposed for the problem of optimally combining classifiers using game-theoretic arguments. Specifically, the problem of designing optimal ensembles of voting classifiers is investigated within the context of Game Theory [2, 3], as an analogy to *n*-person games. A special type of *cooperative games*, namely the *coalition games*, is introduced as the natural setting for formalizing the ensemble design problem, within the scope of Coalition Theory [2, 3] and the Weighted Majority Games (WMG) [2, 3]. This new formulation of the problem leads to the development of a theoretical framework of the weighted voting schemes [1]. Furthermore, this approach leads to *optimal analytical solutions* for the two core problems of: (a) designing the aggregation rule in an optimal way, and (b) assigning optimal voting weights in a voting ensemble of experts. For the problem in (a), the theory of WMG states that the *optimal voting aggregation rules* in a fixed-size ensemble for an arbitrary *n*-label classification task is the weighted majority rule (WMR) [2, 3, 4, 62]; while for the problem in (b), the *optimal voting weights* in such WMR schemes are *calculated analytically* from the experts' competencies, under the conditional independence assumption [4, 5].

This particular type of game-theoretic analytical solution is extremely useful in the process of designing optimal classifier ensembles. The use of simple linear combination models that employ single weights for each classifier, which, however, *do not require iterative training/optimization*, can provide the necessary means to apply multi-classifier schemes in parallel implementations with on-line updating capabilities. In other words, the classifiers can be trained independently and off-line, using any architecture and algorithm available, while *the aggregation scheme involves only direct (analytical) calculation of the voting weight of each classifier*. Another novelty of this paper refers to the notion of the expert's competency, i.e., the prior estimation of the success rate of each individual classifier, as it is required by the WMR optimal rule. The expert's competency is extended to include the *posterior* probabilities associated with each pattern. In practice, the voting weight of each classifier is calculated analytically, in the sense of the WMR formulation, for every sample using the *local accuracy estimates (LAE)*.

---

[1] (MSc, PhD) Associate researcher (post-doc) at Dept. of Informatics & Telecommunications, National Kapodistrian Univ. of Athens (NKUA/UoA), Greece – Email: `harris {at} xgeorgio {dot} info`

[2] (MSc, PhD) Associate professor at Computational Biomedicine Lab., Dept. of Computer Science, University of Houston (UH), Texas, USA – Email: `mmavrof {at} uh {dot} edu`

This paper is organized as follows. Section 2 describes the core aspects of the classification task and its realization under the concept of multi-classifier systems. Section 3 summarizes some basic concepts of Game Theory and phrases the classifier combination task in game-theoretic terms. Section 4 describes the details regarding the datasets and methods used. Section 5 presents the experiments and results. Section 6 is a discussion on the results. Section 7 presents the conclusions.

## 2. Problem statement and current practices

### 2.1 Combining classifiers

The ultimate goal of any pattern recognition system is to design optimally a classifier while at the same time attaining the best generalization performance, for the specific problem at hand. However, even the "best" classifier model can fail on points that other classifiers may succeed in predicting the correct label [6, 1]. Many studies have focused on the possibility of exploiting this complementary nature of the various classifiers, in order to enhance the overall performance. Specifically, each classifier is considered as a trained expert that participates along with others in a "committee", which produces a collective decision according to some well-specified rule.

In the discipline of collective decision-making, a group of $N$ experts, each one with moderate performance levels, are combined in order to produce a collective decision that may be better than the estimate of the best among the experts in the group. According to the famous Condorcet Jury Theorem [7], if the experts' individual decisions are *independent* and their corresponding estimates are more likely to be correct than incorrect ($p_{correct}$>0.5), then an increase in the collective performance, as a group, is guaranteed when the individual estimations are combined. Moreover, this performance continues to grow asymptotically as the size $N$ of the group increases and under the independence assumption. This assertion has been the base for very active experimental and theoretical research in the discipline of pattern recognition.

Over the last decade or so, a wide range of different approaches have been studied to design aggregations or *ensembles* of experts. These employ either a *selection* or a *fusion* scheme [1] to combine the individual classifiers' outputs into a final collective decision. The combining rules vary from very simple to more sophisticated ones. Typical examples include simple averaging and fusion [8, 9], mixture of experts [10, 11], consensus or majority voting [12], dynamic classifier selection [13], supra Bayesian methods [14, 15], evidence-based [16, 17] or template-based [18] decision models. Most of the methods employ weights upon each member in the pool, essentially dictating a corresponding level of confidence to its individual decisions. Hence, the design of such ensembles reduces to the problem of finding these optimal aggregation parameters, with the goal of improving the final accuracy rates of the ensemble.

To comply with the spirit of Condorcet's theorem, a major research effort has been inserted in designing the individual classifiers to be as independent as possible. More recent approaches, such as boosting [19], bagging [20] and random subspace models [21], employ different techniques to increase the level of diversity, essentially by training individual classifiers in different subsets or subspaces of the original set of data.

This requirement can be implemented, in practice, by separating or *splitting* the original training datasets into a new set of distinct or partially overlapping realizations, with respect to: (a) the data samples, (b) the dimensionality, or (c) both. Random subspace methods, most commonly used in aggregation models like the Random Forests [22], are typical examples of using dimensionally-reduced versions of the original data space. Rotation Forests [23] is an example of using both different subsets and different subspaces simultaneously.

## 3. A game-theoretic approach to classifier combination

This section presents a brief overview of the basics of Game Theory, the main concepts of cooperative games and Coalition Theory, as well as the formal definition of weighted majority games (WMG) and weighted majority rules (WMR). Furthermore, the core problem of designing optimal weighted voting schemes for multi-classifier ensembles is introduced within the context of WMG and theoretical analytical solutions are presented under the WMR formulation.

### 3.1 Elements of Game Theory

The mathematical theory of *games* and *gaming* was first developed as a model for situations of conflict. Since the early 1940's, the work of John Von Neumann and Oskar Morgenstern [24] has provided a solid foundation for the most simple types of games, as well as analytical forms for their solutions, with many applications to Economics, Operations Research and Logistics [2, 3]. Each opposing *player* in a game has a set of possible actions to choose from, in the form of *pure* (single choice) or *mixed* (random combination) *strategies*. The set of optimal strategies for all the players is called the *solution* to this game.

    The *zero-sum* games are capable of modeling situations of conflict between two or more players, where one's gain is the other's loss and vice versa. In reality, it is common that in a conflict not all players receive their opponents' looses as their own gain and vice versa. In other words, it is very common a specific combination of decisions among the players to result in a certain amount of loss to one and a corresponding gain, not of equal magnitude, to another. In this case, the game is called *nonzero-sum* and it requires a new set of rules for estimating optimal strategies and solutions.

    During the early 1950's, John Nash has focused primarily on the problem of finding a set of *equilibrium points* in nonzero-sum games, where the players eventually settle after a series of competitive rounds of the game. In 1957 [25], Nash successfully proved that indeed such equilibrium points exist in all nonzero-sum games, defining what is now known as the *Nash theorem* or *Nash solution* to the *bargaining problem* [2, 3]. However, although the Nash theorem ensures that at least one such *Nash equilibrium* exists in all nonzero-sum games, there is no clear indication on how the game's solution can be analytically calculated at this point. In other words, although a solution is known to exist, there is no closed form for nonzero-sum games until today.

    The Nash equilibrium points are not always the globally optimal option for the players. In fact, the Nash equilibrium is optimal only when players are strictly competitive, i.e., when there is no chance for a mutually agreed solution that benefits them more. These strictly competitive forms of games are called *non-cooperative* games. The alternative option, the one that allows communication and prior arrangements between the players, is called a *cooperative* game and it is generally a much more complicated form of nonzero-sum gaming.

### 3.2 Cooperative games and coalition gaming

The problem of cooperative or possibly-cooperative gaming is the most common form of conflict in real life situations. Since nonzero-sum games have at least one equilibrium point, when studied under the strictly competitive form, Nash has comprehensively studied the cooperative option as an extension to it. However, the possibility of finding and mutually adopting a solution that is better for both players than the one suggested by the Nash equilibrium, essentially involves a set of behavioral rules regarding the players' stance and mental state, rather than strict optimality procedures [2, 3]. Nash named this process as *bargain* between the players, trying to mutually agree on one solution between multiple choices within a *bargaining set*. In practice, each player should enter a bargaining procedure if there is a chance that a cooperative solution exists and it provides at least the same gain as the best strictly competitive solution, i.e., the best Nash equilibrium. In this case, if such a solution is agreed between the players, it is called *bargaining solution* of the game. This new framework provides the necessary means to study *n*-person non-cooperative and cooperative games under a unifying point of view. Specifically, a nonzero-sum game can be realized as a strictly competitive or a possibly cooperative form, according to the game's rules and restrictions. Therefore, the cooperative option can be viewed as a generalization to the strictly competitive mode of gaming.

    When players are allowed to cooperate in order to agree on a mutually beneficial solution of game, they essentially choose one strategy over the others and bargain this option with all the others in order to come to an agreement. For *symmetrical* games, i.e., when all players receive the same gains and losses when switching places, this situation is like each player choosing to join a group of other players with similar preference over their initial choice. Each of these groups is called a *coalition* and it constitutes the basic module in this new type of gaming: the members of each coalition act as cooperative players joined together and at the same time each coalition competes over the others in order to impose its own position and become the *winning* coalition. This setup is very common when modeling voting schemes, where the group that captures the relative majority of the votes becomes the winner.

    Coalition Theory [2, 3] is closely related to the classical Game Theory and in particular the cooperative gaming. In essence, each player still tries to maximize its own expectations, not individually any more but instead as part of a greater opposing term. Therefore, the individual gains and capabilities of each player is now

considered in close relation to the coalition this player belongs to, as well as how its individual decision to join or leave a coalition affects this coalition's winning position. The theoretical implications of having competing coalitions of cooperative players, instead of single players, is purely combinatorial in nature, thus making its analysis very complex and cumbersome. There are also special cases of collective decision schemes where a single player is allowed to *abstain* completely from the voting procedure, or prohibit a contrary outcome of the group via a *veto* option. Special sections of Game Theory, namely the *coalition gaming* and *stable sets* in cooperative gaming [2, 3], have studied the effects of introducing "weights" to the choice of each expert according to their competencies, in order to optimize the final decision of the group.

**3.3 Classifier combination as a game-theoretic problem**

The transformation of cooperative *n*-person games into coalition games essentially brings the general problem into a voting situation. Each player casts a vote related to its own choice or strategy, thus constituting him/her as a member of a coalition of players with similar choices. The coalition that gains more votes becomes the winner. In the case where each player selects one out of *M* available options to cast its vote, the collective group decision can be estimated simply by applying the majority voting scheme, i.e., the choice selected is the one gathering the majority of votes. Each subgroup of consentient players essentially represents an opposing assembly to all the other similar subgroups with different consensus of choice.

In the general case where a weight is assigned to each voter and there are *n* available choices to vote for, this form is known in Game Theory as the weighted majority game (WMG) [2, 3]. It has been proven by Nitzan and Paroush (1982) [4] and Shapley and Grofman (1984) [5], that the optimal decision rules for these WMG, in terms of collective performance, are the weighted majority rules (WMR). The same assertion has also been verified independently by Ben-Yashar and Nitzan [26] as the optimal aggregation rule for committees under the scope of informative voting in Decision Theory. This result was later (2001) [62] extended from dichotomous to polychotomous choice situations; hence *the optimality of the WMR formulation has been proven theoretically* for any *n*-label voting task. Furthermore, under the conditional independence assumption, a closed form solution for the voting weights in the WMR formula exists and it is directly linked to each expert's competency. This *optimal weight profile* for the voting experts is the *log of the odds ("log-odds") of their individual competencies* [4, 5].

In this paper, the notion of modeling classification tasks for an ensemble of experts via the precise game-theoretic formulation of WMG and WMR is for the first time applied for combining hard-output (voting) classifiers. Specifically, the design of the combination rule is treated as a standard WMG situation, with each classifier participating in a *simple* coalition game, i.e., choosing the final decision based on the maximum votes (sum of weights) casted. The voting weights in this WMR scheme are calculated in an *analytical* way using the log-odds solution [27, 1]:

$$w_i = \log\left(\frac{p_i}{1-p_i}\right) \quad , \quad p_i = P_i\{\theta = \omega_{correct}\} \quad , \quad i = 1...K \tag{1}$$

where $w_i$ is the combination weight assigned to the *i*-th classifier (player), $p_i$ is the respective estimated probability for correct classification, measured in the validation set, $\theta$ is the predicted class label and $\omega_{correct}$ is the correct class label for **x** (either $\omega_1$ or $\omega_2$), respectively.

Using this game-theoretic analytical solution, the WMR formula is used as the *optimal voting aggregation scheme*, i.e.:

$$O_{wmr}(\mathbf{x}) = \sum_{i=1}^{K} w_i D_i(\mathbf{x}) \tag{2}$$

where $D_i$ is the hard-output of each of the *K* individual classifiers in the ensemble, $w_i$ is its assigned weight, $O_{wmr}$ is the weighted majority sum.

In this study, the classification tasks where chosen to include only dichotomous choice situations, for several reasons (explained later on). Hence, there are only two voting options available (*M*=2) and two class labels to choose from (either $\omega_1$ or $\omega_2$), which essentially simplifies the WMR formulation to a sign-assignment problem:

$$D_{wmr}(\mathbf{x}) = sign\left(O_{wmr}(\mathbf{x}) - T\right) \tag{3}$$

where $D_{wmr}$ is the final decision of the ensemble against a fixed-valued decision threshold $T$, which is typically half the range of values for $O_{wmr}$ [27, 1], i.e.:

$$T = \frac{1}{2}\left(\max_i\{D_i\} + \min_i\{D_i\}\right) \quad , i = 1...K \qquad (4)$$

or simply $T=1/2$ when normalized weights $w_i$ are employed (sum of weights is unity).

Interestingly, although the optimality of this solution under certain conditions has been studied theoretically in the context of many different disciplines, including decision theory and automata theory [28, 29, 30], it is generally considered very limited in terms of optimality, since it does not take into account any dependencies among the trained classifiers on an ensemble.

In this paper, two versions of this WMR-based weighting scheme with respect to the value of $p_i$ are tested: (a) the "static" WMR, using the prior probabilities of correct classification (i.e. "global" competence), and (b) the "adaptive" WMR, using the (estimated) posterior probabilities of correct classification (i.e. "local" competence). In both cases, the success rates are calculated based on a validation set of samples, independently of any training process and any training set used by the classifiers. In this new "adaptive" version of the WMR, which essentially introduces the notion of *local experts* into this framework, the combination weights are calculated so that they reflect the localized (conditional) competencies of the classifiers at each point, i.e.:

$$w_i = \log\left(\frac{p_i}{1-p_i}\right) \quad , \quad p_i = P_i\{\theta = \omega_{correct} \mid \mathbf{x}\} \quad , \quad i = 1...K \qquad (5)$$

This procedure is presented in section 4.5.

## 4. Datasets and Methodology

### 4.1 Selection of benchmark datasets

In order to assess the performance of the various classifier combination methods, publicly available benchmark datasets were considered. Specifically, the Raetch [31, 32, 33] and the ELENA [34] dataset resources were considered and, for the purposes of this study, only 2-class sets with real (non-artificial) data were initially selected. The main reason for employing only dichotomous classification tasks is that multi-class problems essentially add one more layer of complexity in some classification models, especially SVM-based. In practice, the simplification of the classification task itself does not require any second-stage decision, e.g., one-versus-all or pair-wise comparisons. Furthermore, when $M>2$ choices are available in WMG setups, the corresponding WMR decision requires one additional parameter of non-trivial optimization [67], the majority threshold or *quota* ($q$), instead of a simple comparison to the half-sum of the voting weights, as in Eq. (3) and (4). Since the goal of this study is the comparative evaluation of combination rules, while keeping all the other factors as simple as possible, using only 2-class benchmark datasets is a natural choice.

A group of 14 datasets were analyzed in terms of class separability and statistical significance of the corresponding results. In order to make individually trained classifiers as diverse as possible, the method of training them in different subspaces was selected. Consequently, datasets of high dimensionality were preferred. The quantitative criteria used for selecting the final datasets from this group included: (a) The inherent dimensionality of the dataset, in order to be able to use a feature subspace method leading to at least five distinct feature groups. (b) The Chernoff Bound and the corresponding Bhattacharyya Distance [6], as a commonly used class separability measure, when Gaussian distributions are assumed for the classes. (c) Guyon's error counting approach [35] for estimating the minimum size of the test data set, based on the results of a leave-one-out [36] error estimation from a simple OBTC model [37]. (d) The ELENA project's proposal [38] for the minimum number of samples necessary for the estimation of the probability density function (pdf) of a Gaussian probability density, using a probability density kernel estimator, with less than 10% error. (e) The intrinsic dimension was calculated in terms of the fractal dimension estimation method [39, 40] and compared to the real number of distinct features of the dataset, in order to get a quantitative measure of the overall complexity of the sample space and the degree of redundant information among the features. Based on these criteria, the final group of the selected datasets included four candidates from the Raetch packages. These datasets are: 1) Ringnorm, 2) Splice, 3) Twonorm, and 4) Waveform.

**4.2 Dataset split and subspace method**

Each base dataset was randomly separated into a base training set and a test set of samples. Diversity among the classifiers was introduced in the ensemble by training them in different subspaces. Random subspace methods [21] have been successfully used in the past as the means to increase diversity among classifiers. However, in these methods, the grouping of distinct features into subsets is conducted randomly and involves either distinct or overlapping memberships of features in the various groups. In contrast, in this study a non-random subspace procedure was implemented by using a feature ranking method and a subsequent grouping into distinct subsets, in order to achieve more or less equal discrimination power. A much simpler version of this method has been used successfully in the past [41]. These approaches are generally referred to as "ranked" subspace methods, since subsets of features are evaluated and ranked according to some specific statistical criterion, in order to control the discrimination power and the robustness of each subspace in various classification or clustering applications (see e.g. [62]).

In this paper, the training set was partitioned into $K$ distinct feature groups. Each group of features was created in a way that satisfied three basic constraints: (a) each group to be distinct, i.e., no feature is common in any two groups, (b) all the features are used "as-is", i.e., no projection or other complex transformation is applied (e.g. PCA), and (c) each feature group to represent approximately the same class-discrimination potential. The third constraint requires a method for ranking all the features in terms of discrimination power, against the two classes, as well as their statistical independence with regard to the other features in the initial training set. The MANOVA method [42] was used to assign a multivariate statistical significance value to each one of the features and then produce a sorted list, based on (the log of) this value.

A "fair" partitioning of this sorted list of features into equally "accurate" groups, in terms of classification results, was conducted by selecting features in pairs from the top and bottom positions, assigning the currently "best" and "worst" features in the same group. Furthermore, the efficiency of each group was estimated in terms of summing the log of the statistical significance value, assigned by MANOVA, of all the features contained in this group. The log was employed in order to avoid excessive differences between the values assigned by MANOVA, thus creating more even subset sums of these values. Practically, every such pair of features was assigned in groups sequentially, in a way that all groups contained features with approximately equal sum of the log of the values assigned by MANOVA.

Each one of these $K$ distinct feature groups was used for training one of the $K$ classifiers in the ensemble. As a result, the issue of the desired diversity between the classifiers of the ensemble is addressed independently from the combination rules themselves, making their subsequent comparison easier and more realistic. It should be noted, that the goal of this study is not classifier independence or diversity, but rather to evaluate the performance of the WMR and other combination rules, using *weakly* independent classifiers, i.e., without guaranteed diversity. In fact, the introduction of a feature subspace method only creates *some* diversity, which makes the evaluation of the ensembles more realistic.

**4.3 Classifier models**

Three, among the most popular, classifier models were selected to form committees of experts, in order to test the various classifier combination schemes. Specifically, the Support Vector Machine (SVM) [43, 44], the (weighted) k-nearest-neighbor (w/$k$-NN) [6] and the Decision Tree (DT) [45] classifiers were employed in this study.

For the SVM architecture, a geometric nearest point algorithm (NPA) [46], based on the notion of reduced convex hulls (RCH) [47], was used for training a standard SVM architecture with radial-basis function (RBF) as the kernel of the non-linear mapping.

The tree-based classification models were selected as a very typical candidate of unstable classifiers, already used successfully in other combination schemes. In their simplest form, each tree node contains a threshold value that is compared to one of the input features and the result dictates which of two possible paths to follow towards the next tree level. This type of decision trees is often referred to as Ordinary Binary Classification Trees (OBCT) [6]. In this study, the classic Classification and Regression Tree (CART) algorithm [45] was employed for designing soft-output (regression) and hard-output (classification) DT, used in conjunction with soft- and hard-output combination rules, respectively. Three splitting criteria were tested separately: (a) the *Gini index* of diversity, a typical choice in CART models that is similar to entropy, (b) the *twoing* criterion, optimizing the criterion of splitting the contents of each node into two disjoint and mutually exclusive subsets, and (c) the *deviance* criterion, which maximizes the variability (variance) reduction within each of the two splits of the node. These three splitting criteria, which are some of the most commonly used choices in typical OBCT models, were tested separately for completeness purposes.

Finally, a modified version of the w/$k$-NN classifier was employed [48, 49, 50]. The typical $k$-NN classifier architecture was enriched with the options of choosing distance functions other than the classic Euclidean, employing a non-constant weighting function to the test samples around the center of the $k$-closure neighborhood. In this study, the weighting functions were fixed (non-trained), symmetric around the center of the $k$-neighborhood and scaled appropriately. In each case, the smallest weight value was assigned to the furthest of the $k$ neighbors and the largest weight value to the center of the $k$-neighborhood. In other words, the weighting profile was either constant, in the case of the typical non-weighted $k$-NN implementation, or a constantly decreasing function around the center of the $k$-neighborhood. The distance metrics implemented for this w/$k$-NN classifier were the Euclidean, the city block, the Minkowski, the cosine, the correlation, the Mahalanobis, the Chebychev, and the Hamming kernels [6]. The weighting metrics implemented for this w/$k$-NN classifier were the constant (classic, no weighting), the linearly decreasing and the Gaussian profiles. The introduction of the different distance functions and especially the option of employing weights to the $k$-neighbors according to their distance from the center of the local test set, had little effect to the overall performance of the w/$k$-NN classifier but produced much more stable soft-output profiles, which were used subsequently for the calculation of local accuracy estimates (see: section 4.5).

### 4.4 Combination rules

A total of eight combination rules were examined in this study. Specifically, four typical hard-output combination methods were employed (namely one classic rank-based method and three voting-based schemes, including the "static" and "adaptive" versions of WMR), two soft-output averaging methods and two Bayesian-based combination rules.

The standard maximum rule was employed as a typical rank-based method [1, 51]:

- maximum ("STD: maximum"):

$$O_{\max}(\mathbf{x}) = \omega_s : p_s = \max_{j=1\ldots 2}\left\{\max_{i=1\ldots K}\{\mu_{ij}\}\right\},\ \mu_{ij} = O_i(\mathbf{x})\,|\,\bigl(\theta(\mathbf{x}) = \omega_j\bigr) \quad (6)$$

where: $\mathbf{x}$ is the current input sample to be classified, $O_i(\mathbf{x})$ is output value by the $i$-th classifier for class label $\omega_j$ given $\mathbf{x}$, $\theta(\mathbf{x})$ is the predicted class label, $\{\omega_1, \omega_2\}$ are the two class labels, $K$ is the size of the ensemble, $\mu_{ij}$ is the corresponding support value by the $i$-th classifier for class label $\omega_j$ given $\mathbf{x}$, and $p_s$ is the selected support value.

It should be noted that two other typical rank-based combination rules are equivalent to the maximum and the simple majority rules, respectively, in case of dichotomous choice classification [1]. Specifically, the class labels selected by the minimum rule [51] are the same to the ones selected by the corresponding maximum rule that uses the same support values. Similarly, the class labels selected by the median rule [51] are the same to the ones selected by the corresponding simple majority rule that uses the same support values.

- simple majority voting ("STD: simple majority") [1, 52]:

$$O_{maj}(\mathbf{x}) = \sum_{i=1}^{K} D_i(\mathbf{x}) \quad (7)$$

$$D_{maj}(\mathbf{x}) = sign\bigl(O_{maj}(\mathbf{x}) - T\bigr) \quad (8)$$

where $D_i$ is the hard-output of each of the $K$ individual classifiers in the ensemble, $O_{maj}$ is the majority sum. The final hard-output decision $D_{maj}$ of the simple majority rule is calculated against a fixed-valued decision threshold $T$, which is typically half the range of values for $O_{maj}$ [1]:

$$T = \tfrac{1}{2}\left(\max_i\{D_i\} + \min_i\{D_i\}\right),\ i = 1\ldots K \quad (9)$$

which is the same to the one employed for WMR (in Eq.4) but with $w_i = 1/K$ for the simple majority rule.

Additionally, two soft-output averaging models were included, a non-weighted and a weighted one [1]:

- simple average ("STD: simple average"):

$$O_{avg}(\mathbf{x}) = \sum_{i=1}^{K} w_i O_i(\mathbf{x}) \quad , w_i = 1/K \quad (10)$$

- weighted average ("STD: LSE-weighted average"):

$$O_{lsewavg}(\mathbf{x}) = \sum_{i=1}^{K} \hat{w}_i O_i(\mathbf{x}) \quad (11)$$

where $O_i(\mathbf{x})$, $O_{avg}(\mathbf{x})$ and $O_{lsewavg}(\mathbf{x})$ are the soft-output value of the single classifier (ensemble member), the simple averaging rule and the weighted average rule, respectively. For the simple average rule, all weights are equal, i.e., $w_i=1/K$. The vector $\hat{\mathbf{w}}$ is the optimal one for the weighted average rule, estimated by a linear regression formula on the individual classifier outputs, against the correct classification tag, in terms of a least-squares error (LSE) minimization criterion [53, 1]. Thus, this method can be considered as an example of a "trained" linear weighting rule of soft-output classifiers. In contrast, *the WMR approach employs fixed analytical optimal weighting profile* and hard-output classifications (votes) as input, *with no need for further training*.

Finally, two Bayesian-based combination rules were employed as a very efficient and simple implementation of non-weighted schemes, which exploit information about local accuracy estimates. Specifically, the method of Dynamic Classifier Selection based on Local Accuracy (DCS-LA) [13, 54, 55] was employed as a typical example of a local accuracy-based non-weighted combination rule, using the notion of overall local accuracy [13]. Two different variants of this model (DCS-LA variants) were implemented:

- employing the full Bayes rule for the conditional probabilities ("STD: DCS-LA (with priors)"):

$$O_{bayesP}(\mathbf{x}) = D_s(\mathbf{x}) : p_s = \max_{i=1...K}\{p_i\} \quad , p_i = P_i\{\theta = \omega_{correct} | \mathbf{x}\} \cdot P_i\{\theta = \omega_{correct}\} \quad (12)$$

- or, using only the local accuracy estimate itself ("STD: DCS-LA (no priors)"):

$$O_{bayes}(\mathbf{x}) = D_s(\mathbf{x}) : p_s = \max_{i=1...K}\{p_i\} \quad , p_i = P_i\{\theta = \omega_{correct} | \mathbf{x}\} \quad (13)$$

The final decision in these types of DCS-LA models is dictated by the expert with the highest conditional probability of success, i.e., highest "confidence". As a result, the model implemented in this study is essentially a direct implementation of the standard Bayes decision theory that is based on maximizing the likelihood of "correct" classification for the current input data $\mathbf{x}$. It should be noted that, although the Bayes rule includes division by the pdf against the input data $\mathbf{x}$, this factor is irrelevant here since the model is applied to a specific input sample and therefore the corresponding pdf value is always equal to unity.

For all the soft-output combination rules (simple and LSE-weighted averaging, DCS-LA models), the final decision is calculated against a fixed threshold value similarly to the majority voting rules, which is typically half the range of values for the specific combination rule, i.e.:

$$D_{combR}(\mathbf{x}) = sign(O_{combR}(\mathbf{x}) - T) \quad (14)$$

$$T = \frac{1}{2}\left(\max_i\{O_i\} + \min_i\{O_i\}\right) \quad , i=1...K \quad (15)$$

where $O_i$ is the soft-output of each of the $K$ individual classifiers in the ensemble, $O_{combR}$ is the soft-output of the combination rule and $D_{combR}$ is the final class decision (thresholded value).

Table 1 summarizes all the eight combination rules used in this study.

**Table 1:** Overview of the eight combination rules used in this study.

|  | Non-weighted | Weighted |
|---|---|---|
| **Static or Rank-based** | simple average | LSE-weighted average |
|  | maximum | WMR (static) logodds |
|  | majority |  |
| **Adaptive (using posteriors)** | Bayesian (DCS-LA) no priors | WMR (adaptive) logodds |
|  | Bayesian (DCS-LA) with priors |  |

**4.5 Local accuracy estimates method**

To compute the local accuracy estimates, as required by the DCS-LA method, as well as the modification of the WMR as pointed out in section 3.3, the "overall" local accuracy method was adopted [13, 55]. To this end, we chose to estimate the error pdf directly by employing a non-parametric density-based method, by means of histogram approximation [6]. For one-dimensional probability functions, the histogram method has been proven more efficient than direct interpolation through isotonic regression functions [56]. In order to avoid non-uniformities in the distribution of the classifiers' soft-output values, the use of *dynamic bin width allocation* [57] was employed instead of equal bin width. This method ensures that every bin contains roughly the same number of samples, i.e., the width of the histogram bins is adjusted appropriately, in order to produce uniform resolution and smoothness of the histogram curve throughout the entire range of values.

In this study, the local accuracy estimation of each classifier was based on the corresponding error pdf, approximated via the histogram method with dynamic bin width allocation. The complete process includes five distinct steps: (a) Each classifier in the ensemble is designed based on a training set of samples and subsequently evaluated on a different set of test samples. (b) The soft-output values of the classifier are then distributed uniformly into bins of dynamic width. (c) The corresponding localized error probability estimation is calculated for every bin in terms of error frequency ratio (errors versus total samples in bin). (d) After the calculation of the error pdf value in each bin, the bin values are interpolated with a piecewise cubic Hermite spline, chosen for its shape-preserving properties [58, 59], in order to produce the final, continuous, error pdf estimator. (e) The local accuracy estimate, i.e., the "success" pdf, for each specific classifier is calculated by one minus its corresponding error pdf (interpolated) value for a given input sample **x**, i.e.:

$$P_i\{\theta = \omega_{error} | \mathbf{x}\} = Ne_i^m / N_i^m \quad , b_i^m \geq O_i(\mathbf{x}) > b_i^{m+1} \quad (16)$$

$$LAE: P_i\{\theta = \omega_{correct} | \mathbf{x}\} = 1 - H_i^m\left(P_i\{\theta = \omega_{error} | \mathbf{x}\}\right) \quad , b_i^m \geq O_i(\mathbf{x}) > b_i^{m+1} \quad (17)$$

where $O_i(\mathbf{x})$ is the $i$-th classifier's output, $N_i^m$ is the number of output values from the $i$-th classifier in (dynamic) histogram bin $m$, $Ne_i^m$ is the number of incorrect classifications committed by the $i$-th classifier in bin $m$, $b_i^m$ and $b_i^{m+1}$ are the boundaries of bin $m$ for the $i$-th classifier, and $H_i^m$ is the piecewise cubic Hermite spline for the error pdf in bin $m$ for the $i$-th classifier.

Results from all single-classifier tests have shown that the local accuracy estimation based on the dynamic-width histogram method produced very robust and accurate results for all classifiers, even for the OBTC, which is the most unstable of the three base classifiers used in this study. Figure 1 illustrates a real example of the procedure. The plot in (a) illustrates the fixed-width histogram (counts per fixed-width bin) of the SVM kernel output values against correct (green/high) and incorrect (red/low) classifications, for the single-classifier configuration in the "splice" dataset. The plot in (b) illustrates the resulting dynamic-width normalized histogram (dots) and the corresponding local accuracy estimator (interpolated dynamic-width bins) function, which calculates the estimated posterior probability of successful (blue/high) and incorrect (red/low) classification with regard to the current SVM kernel output.

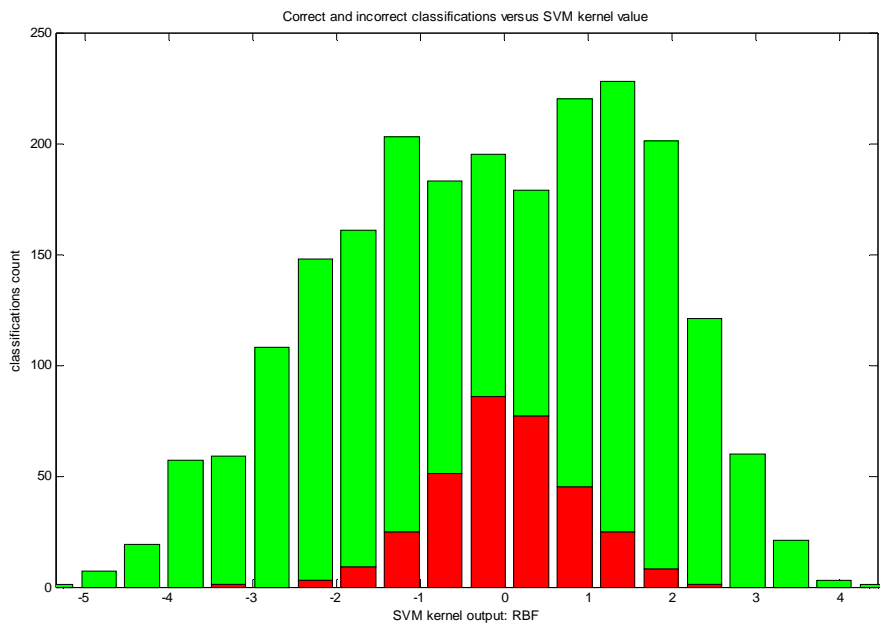

(a)

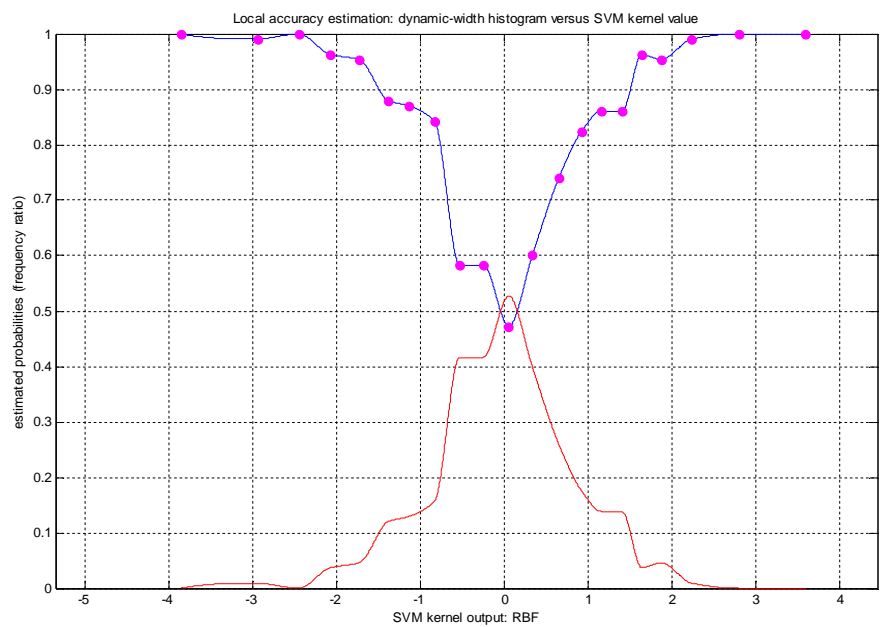

(b)

**Figure 1:** Example of the local accuracy estimation method. Green (high/lighter) and red (low/darker) portions of the bars in the fixed-width histogram in plot (a) represent number of correct and incorrect classifications, respectively, with regard to the classifier's output value. The marked points in plot (b) correspond to the new bin centers of the dynamic-width histogram for the "success" pdf, while the interpolation curve represents the analytical local accuracy estimator function for correct (blue/high/darker curve) and incorrect (red/low/lighter curve) classification, with regard to the classifier's output value. The results are for the single SVM classifier on the "splice" dataset.

## 5. Experiments and Results

The experimentation phase of this study involved three consecutive stages: (a) the characterization of all datasets based on single classifier tests, (b) the training of ensembles based on different choices of datasets, number and type of base classifiers, and (c) the comparative evaluation of all the classifier combination rules.

### 5.1 Datasets and classifiers characterization

In the preliminary analysis stage, every dataset/classifier combination involved five training realizations, each one employing training and parameter optimization according to each classifier model. For the SVM classifier, the optimization included the $\mu$ soft margin parameter of the NPA-RCH training algorithm [46, 47], as well as the *epsilon* parameter (convergence accuracy). Furthermore, the Radial Basis Function (RBF) kernel [44] with optimized *sigma* ($\sigma$) value was used in the final choice of the SVM classifiers' structure. For the OBTC, each setup included a classification and a regression tree, both trained on the same data and both using the same splitting criterion (one of: *Gini*, *twoing* or *deviance*). In all cases, the OBTC models were optimized against the exact choice of the splitting function and they all employed a minimum limiting threshold of ten samples per splitting node during training/pruning phases [1, 6]. Finally, for the w/$k$-NN classifier, the optimization included the best choice for distance function (Euclidean or other), the $k$-size parameter and the weighting function (constant or other).

Tables 2 and 3 illustrate all the dataset/classifier combinations and the corresponding best-accuracy configurations for the full feature sets (i.e., no feature selection/optimization), which were used as the base for creating and evaluating the corresponding ensembles in the subsequent stage of the experiments.

**Table 2:** Dataset specifications and single-classifier (reference) accuracies (%). Mean and standard deviation values are based on five training realizations with full feature set.

| Dataset | Training set size | Testing set size | Dataset Dimension | SVM accuracy | OBTC accuracy | w/$k$-NN accuracy |
|---|---|---|---|---|---|---|
| ringnorn | 400 | 7000 | 20 | **97.66 ± 0.22** | 80.78 ± 2.41 | 77.00 ± 1.83 |
| splice | 1000 | 2175 | 60 | 85.29 ± 1.08 | **92.98 ± 0.97** | 77.88 ± 2.09 |
| twonorm | 400 | 7000 | 20 | 97.70 ± 0.15 | 76.56 ± 1.63 | **97.85 ± 1.52** |
| waveform | 400 | 4600 | 21 | **90.10 ± 0.40** | 80.91 ± 1.65 | 89.85 ± 1.50 |

**Table 3:** Single-classifier best configurations against datasets, based on five training realizations.

| Dataset | best SVM configuration | best OBTC configuration | best w/$k$-NN configuration |
|---|---|---|---|
| ringnorn | kernel: RBF ($\sigma$=5) <br> $\mu$=0.016 <br> *epsilon*=5.e-4 | splt.func=*twoing* <br> splt.limit=10 | dist.func=*Euclidean* <br> $k$-size=1 <br> weight.func=*none* |
| splice | kernel: RBF ($\sigma$=42) <br> $\mu$=0.036 <br> *epsilon*=5.e-4 | splt.func=*deviance* <br> splt.limit=10 | dist.func=*Euclidean* <br> $k$-size=15 <br> weight.func=*Gaussian* |
| twonorm | kernel: RBF ($\sigma$=100) <br> $\mu$=0.008 <br> *epsilon*=5.e-4 | splt.func=*deviance* <br> splt.limit=10 | dist.func=*Euclidean* <br> $k$-size=17 <br> weight.func=*linear* |
| waveform | kernel: RBF ($\sigma$=20) <br> $\mu$=0.020 <br> *epsilon*=5.e-4 | splt.func=*Gini* <br> splt.limit=10 | dist.func=*Euclidean* <br> $k$-size=21 <br> weight.func=*none* |

### 5.2 Classifier ensembles training

Using the best-configuration results obtained from the previous experimentation stage, two different base ensemble designs were employed in the second stage. Specifically, the feature subspace method was used to create dataset splits for $K=5$ and $K=7$ feature subsets, with the application of MANOVA as the feature ranking method for the complete (original) datasets. Each of these $K$-splits setups was applied to create ensembles with five or seven classifiers, each employing one of the three base types of classifiers (SVM, OBTC, w/$k$-NN). Subsequently, for each one of these ensemble setups, ten random realizations of training and testing subsets of the dataset were created, using the training/testing ratio also used for the corresponding single-classifier cases (see: Table 2). This procedure was employed in the same way for all four base datasets, for ensembles of five or seven (same type) classifiers.

For the training of the classifiers in any given ensemble setup, the model parameters used were the same with the ones calculated in the corresponding single-classifier case during the first experimentation phase. The main reason for not employing a full optimization procedure in this second stage was the fact that the subsequent comparison of the different combination rules was based in their *relative* differences in performance between them and not with the corresponding single classifier model (using the full feature set). Moreover, this procedure of optimizing every individual classifier in the ensemble would result in the increase of the total processing time with no actual benefit to the purposes of this particular study, since the goal here is to test the efficiency and robustness of the various combination rules in ensembles of weak or sub-optimally trained classifiers.

Table 4 illustrates the mean and standard deviation values of average and maximum (in parentheses) accuracy rates of ensemble members, for different choices of datasets, $K$-splits and base classifiers.

**Table 4:** Average classifier group accuracies (%) for all datasets and splits, against all the ensemble members in ten training realizations. The numbers enclosed in parentheses indicate the mean of the maximum-accuracy members in each corresponding ensemble, i.e., the average over only the top members across the ten training realizations.

| | | Classifiers | | |
|---|---|---|---|---|
| **Dataset** | ***K*-splits** | **SVM** | **OBTC** | **w/*k*-NN** |
| ringnorm | 5 | 77.63 ± 2.13 *(81.10 ± 1.05)* | 77.42 ± 0.69 *(79.29 ± 0.68)* | 73.64 ± 0.55 *(74.85 ± 0.52)* |
| ringnorm | 7 | 72.89 ± 1.04 *(80.53 ± 1.73)* | 74.30 ± 1.04 *(78.21 ± 1.25)* | 69.65 ± 0.43 *(73.43 ± 0.91)* |
| splice | 5 | 66.80 ± 0.98 *(78.94 ± 0.50)* | 72.57 ± 0.46 *(88.39 ± 0.33)* | 67.33 ± 0.78 *(80.84 ± 0.71)* |
| splice | 7 | 63.19 ± 1.30 *(77.17 ± 1.37)* | 68.70 ± 0.48 *(82.49 ± 0.49)* | 66.04 ± 0.34 *(80.32 ± 0.41)* |
| twonorm | 5 | 81.11 ± 0.11 *(84.11 ± 0.14)* | 73.82 ± 1.30 *(75.34 ± 0.93)* | 79.87 ± 0.16 *(82.72 ± 0.34)* |
| twonorm | 7 | 76.83 ± 0.65 *(82.09 ± 0.65)* | 71.78 ± 0.44 *(75.26 ± 0.76)* | 75.73 ± 0.20 *(80.40 ± 0.51)* |
| waveform | 5 | 75.72 ± 1.28 *(79.26 ± 0.59)* | 78.18 ± 0.63 *(79.57 ± 0.62)* | 79.92 ± 0.48 *(81.43 ± 0.53)* |
| waveform | 7 | 70.25 ± 1.09 *(78.54 ± 1.03)* | 76.65 ± 0.49 *(79.40 ± 0.46)* | 77.64 ± 0.38 *(80.85 ± 0.45)* |

With respect to the dataset split process employed in this study, Table 4 demonstrates that the non-random (MANOVA ranking) feature subspace method produced more or less balanced ensembles, as the standard deviation on the mean and maximum single-member accuracies remained very low in all cases, regardless of the type of base classifier employed.

**5.3 Testing of combination rules in the ensembles**

In each training/testing cycle, the classification outputs from the pool of *K* classifiers were fed as input to each of the eight combination schemes (discussed in section 4.4) investigated in this study, producing the corresponding combined classification outputs of the ensemble.

It should be noted that the half-range decision threshold was used in all combination rules, i.e., no analytical optimization was conducted for *T*. This choice is justified by the results from previous studies [e.g., 41, 1], which support the assertion that the optimized *T* value rarely lies far from the half-range value. Furthermore, in the case of combination rules that employ local accuracy estimates (i.e., a pdf approximation), information about the shape and properties of the decision boundary of each classifier is already encoded partially in the (estimated) conditional probability of "correct" classification, and is used either directly (in the case of DCS-LA) or indirectly (in the weights of WMR).

In the sequel, the overall relative performance of each combination rule was determined in terms of ranking position for each case, i.e., according to its corresponding improvement over the mean group accuracy, for each dataset and *K* value employed. Specifically, a *weighted Borda* or *w/Borda* count method [60] was employed to attribute ten points to the top-ranked combination rule (first on the list of eight rules), nine points to the second (second on the list of eight rules), and so on. In case of a "tie" where two combination rules exhibited exactly the same performance, both got the same w/Borda points for the specific ranking position. The Borda and w/Borda count methods are often used in cases when an overall evaluation of classifiers or ensembles is required over a wide range of different configurations, datasets and average "grouped" performances, i.e., when direct aggregation of individual "group" success rates is not valid in terms of statistical context. In this study, as the average accuracies of all the models are more or less compact within-datasets but very different across-datasets (very different classification tasks), their *relative* ranking is much more informative than mean and standard deviation calculations of actual accuracy rates.

Table 5 illustrates the wBorda rankings of each combination rule, as well as the mean increase in accuracy (each cell corresponds to the average over ten training/testing realizations) over the mean accuracy of the individual members in the ensemble of SVM classifiers, for each dataset and *K*-split value employed.

**Table 5:** SVM ensemble results for all combination rules, datasets and *K*-splits. Improvements on average group accuracy (%) are presented in decimal numbers, while wBorda ranking values are presented as integers. The underlined wBorda values indicate top-ranking positions (10 points). All accuracy improvements were calculated as the difference of accuracy rates between the ensemble and the corresponding single-classifier configuration. Negative values indicate deterioration in performance. The combination rules are presented sorted against the SUM column, which represents the total sum of wBorda points assigned to each combination rule over all datasets and *K*-splits.

|  | ringnorm | | splice | | twonorm | | waveform | | SUM | MEAN | STDEV |
| --- | --- | --- | --- | --- | --- | --- | --- | --- | --- | --- | --- |
|  | *K*=5 | *K*=7 | *K*=5 | *K*=7 | *K*=5 | *K*=7 | *K*=5 | *K*=7 | | | |
| STD: DCS-LA (no priors) | 8 | 8 | <u>10</u> | <u>10</u> | 9 | 6 | 9 | <u>10</u> | **70** | 8.75 | 1.39 |
|  | 16.96 | 20.25 | 19.69 | 22.04 | 14.55 | 17.28 | 8.06 | 12.87 | | 16.46 | 4.53 |
| WMR: logodds (adaptive) | 7 | 6 | 8 | 8 | 8 | <u>10</u> | <u>10</u> | 9 | **66** | 8.25 | 1.39 |
|  | 16.11 | 18.11 | 17.68 | 20.06 | 14.22 | 18.80 | 8.67 | 12.8 | | 15.81 | 3.75 |
| STD: DCS-LA (w/priors) | 9 | 9 | 9 | 9 | 9 | 5 | 4 | 3 | **57** | 7.13 | 2.64 |
|  | 17.27 | 20.43 | 19.17 | 21.89 | 14.55 | 17.22 | -3.31 | 4.25 | | 13.93 | 8.83 |
| STD: simple average | <u>10</u> | <u>10</u> | 4 | 4 | <u>10</u> | 4 | 7 | 8 | **57** | 7.13 | 2.80 |
|  | 17.64 | 21.62 | 7.76 | 6.11 | 15.20 | 5.82 | 7.97 | 11.68 | | 11.73 | 5.87 |
| STD: LSE-w/average | 6 | 5 | 7 | 6 | 7 | 7 | 8 | 7 | **53** | 6.63 | 0.92 |
|  | 14.81 | 15.58 | 16.46 | 19.04 | 13.76 | 17.49 | 7.98 | 11.21 | | 14.54 | 3.56 |
| WMR: logodds (static) | 4 | 4 | 6 | 7 | 7 | 9 | 6 | 5 | **48** | 6.00 | 1.69 |
|  | 14.63 | 13.62 | 16.41 | 19.17 | 13.76 | 17.66 | 7.37 | 9.77 | | 14.05 | 3.93 |
| STD: simple majority | 5 | 3 | 5 | 5 | 7 | 8 | 6 | 4 | **43** | 5.38 | 1.60 |
|  | 14.81 | 12.57 | 12.28 | 13.92 | 13.76 | 17.64 | 7.37 | 7.34 | | 12.46 | 3.55 |
| STD: maximum | 3 | 7 | 3 | 3 | 6 | 3 | 5 | 6 | **36** | 4.50 | 1.69 |
|  | 14.41 | 18.68 | 1.68 | -2.18 | 12.10 | 1.64 | 6.07 | 10.16 | | 7.82 | 7.21 |

The same approach was applied to ensembles of OBTC and w/*k*-NN classifiers. Tables 6 and 7 illustrate the wBorda rankings and mean accuracy improvements of OBTC and w/*k*-NN ensembles, accordingly.

**Table 6:** OBTC ensemble results for all combination rules, datasets and *K*-splits. The adverted notation is the same as in Table 5.

|  | ringnorm | | splice | | twonorm | | waveform | | SUM | MEAN | STDEV |
|---|---|---|---|---|---|---|---|---|---|---|---|
|  | *K*=5 | *K*=7 | *K*=5 | *K*=7 | *K*=5 | *K*=7 | *K*=5 | *K*=7 | | | |
| WMR: logodds (adaptive) | <u>10</u> | 8 | 6 | 7 | 9 | <u>10</u> | <u>10</u> | <u>10</u> | **70** | 8.75 | 1.58 |
|  | 14.25 | 15.34 | 18.93 | 21.49 | 15.00 | 18.44 | 7.46 | 9.78 | | 15.09 | 4.69 |
| WMR: logodds (static) | 9 | 7 | 8 | 8 | 8 | 8 | 9 | 9 | **66** | 8.25 | 0.71 |
|  | 14.17 | 14.88 | 19.16 | 21.58 | 14.40 | 17.84 | 7.10 | 9.23 | | 14.76 | 4.86 |
| STD: LSE-w/average | 9 | 9 | 7 | 9 | 8 | 7 | 8 | 8 | **65** | 8.13 | 0.83 |
|  | 14.17 | 15.40 | 18.98 | 21.65 | 14.40 | 17.83 | 7.05 | 9.17 | | 14.83 | 4.87 |
| STD: simple majority | 9 | 6 | 5 | 5 | 8 | 9 | 9 | 9 | **60** | 7.50 | 1.85 |
|  | 14.17 | 14.86 | 15.14 | 16.26 | 14.40 | 17.87 | 7.10 | 9.23 | | 13.63 | 3.62 |
| STD: simple average | 8 | <u>10</u> | 4 | 4 | <u>10</u> | 6 | 7 | 7 | **56** | 7.00 | 2.33 |
|  | 13.51 | 16.07 | 12.14 | 15.10 | 15.32 | 16.66 | 5.49 | 6.60 | | 12.61 | 4.31 |
| STD: DCS-LA (no priors) | 7 | 4 | <u>10</u> | <u>10</u> | 7 | 4 | 6 | 6 | **54** | 6.75 | 2.31 |
|  | 9.23 | 11.00 | 20.67 | 21.92 | 13.34 | 12.95 | 3.89 | 4.02 | | 12.13 | 6.70 |
| STD: DCS-LA (w/priors) | 6 | 5 | 9 | 6 | 6 | 5 | 4 | 4 | **45** | 5.63 | 1.60 |
|  | 8.34 | 11.18 | 20.57 | 20.90 | 13.33 | 12.99 | -5.00 | -6.71 | | 9.45 | 10.40 |
| STD: maximum | 5 | 3 | 3 | 3 | 5 | 3 | 5 | 5 | **32** | 4.00 | 1.07 |
|  | 0.52 | -1.48 | 8.13 | 9.25 | 2.05 | -1.58 | -1.37 | -3.46 | | 1.51 | 4.73 |

**Table 7:** w/*k*-NN ensemble results for all combination rules, datasets and *K*-splits. The adverted notation is the same as in Table 5.

|  | ringnorm | | splice | | twonorm | | waveform | | SUM | MEAN | STDEV |
|---|---|---|---|---|---|---|---|---|---|---|---|
|  | *K*=5 | *K*=7 | *K*=5 | *K*=7 | *K*=5 | *K*=7 | *K*=5 | *K*=7 | | | |
| WMR: logodds (adaptive) | 8 | 8 | 8 | 7 | <u>10</u> | <u>10</u> | <u>10</u> | <u>10</u> | **71** | 8.88 | 1.25 |
|  | 17.17 | 20.76 | 16.22 | 19.03 | 14.16 | 18.08 | 6.98 | 9.12 | | 15.19 | 4.85 |
| STD: DCS-LA (no priors) | 9 | 9 | 9 | <u>10</u> | 7 | 6 | 6 | 6 | **62** | 7.75 | 1.67 |
|  | 19.68 | 21.41 | 19.05 | 21.49 | 9.48 | 10.66 | 4.55 | 3.65 | | 13.75 | 7.52 |
| STD: DCS-LA (w/priors) | <u>10</u> | <u>10</u> | <u>10</u> | 9 | 6 | 5 | 4 | 4 | **58** | 7.25 | 2.76 |
|  | 20.00 | 21.79 | 19.30 | 21.24 | 9.45 | 10.64 | -7.51 | -8.96 | | 10.74 | 12.62 |
| STD: LSE-w/average | 7 | 5 | 5 | 5 | 9 | 8 | 9 | 9 | **57** | 7.13 | 1.89 |
|  | 13.47 | 16.99 | 13.91 | 17.34 | 14.08 | 17.85 | 6.77 | 8.86 | | 13.66 | 4.02 |
| WMR: logodds (static) | 7 | 6 | 4 | 4 | 9 | 9 | 8 | 8 | **55** | 6.88 | 2.03 |
|  | 13.47 | 17.02 | 13.78 | 17.09 | 14.08 | 17.93 | 6.76 | 8.82 | | 13.62 | 4.01 |
| STD: simple average | 6 | 4 | 7 | 8 | 8 | 7 | 7 | 7 | **54** | 6.75 | 1.28 |
|  | 3.03 | -1.72 | 15.41 | 19.20 | 12.82 | 13.75 | 6.42 | 6.16 | | 9.38 | 7.03 |
| STD: simple majority | 7 | 7 | 3 | 3 | 9 | 9 | 8 | 8 | **54** | 6.75 | 2.43 |
|  | 13.47 | 17.03 | 9.39 | 12.73 | 14.08 | 17.93 | 6.76 | 8.82 | | 12.53 | 3.96 |
| STD: maximum | 5 | 3 | 6 | 6 | 6 | 3 | 5 | 5 | **39** | 4.88 | 1.25 |
|  | -8.01 | -9.72 | 15.29 | 18.44 | 5.82 | 2.07 | 4.07 | 2.07 | | 3.75 | 9.85 |

Table 8 presents a summary of the wBorda rankings from Tables 5, 6 and 7. The list of all the combination rules is sorted according to their sum of wBorda points, i.e., their overall efficiency throughout all the base datasets (four) and *K*-splits (*K*=5, *K*=7). Additionally, based on the results from Tables 2 and 4 through 7, Table 9 presents a summary of the comparative performance of the best ensemble designs against the corresponding best single-classifier performance, for all datasets.

**Table 8:** Overall evaluation of all the combination rules, using the wBorda results from all the experiments. The wBorda sum, mean and standard deviation values for each combination rule were calculated across all the datasets, *K*-splits and classifiers. The list is sorted according to the wBorda sum (and mean) ranking position of each combination rule, from the best to the worst combination rule.

| Combination rule | w/Borda SUM | w/Borda MEAN | w/Borda STDEV |
|---|---|---|---|
| WMR: logodds (adaptive) | 207 | 8.63 | 1.38 |
| STD: LSE-w/average | 175 | 7.29 | 1.40 |
| STD: DCS-LA (no priors) | 173 | 7.21 | 2.19 |
| WMR: logodds (static) | 169 | 7.04 | 1.78 |
| STD: simple average | 167 | 6.96 | 2.14 |
| STD: DCS-LA (w/priors) | 160 | 6.67 | 2.41 |
| STD: simple majority | 157 | 6.54 | 2.11 |
| STD: maximum | 107 | 4.46 | 1.35 |

**Table 9:** Overall evaluation of the best ensemble designs against the best single-classifier configurations, for all datasets. The values in the rightmost column refer to the difference between the accuracy (%) of the best ensemble design and the corresponding best single-classifier accuracy for a specific dataset.

| Dataset | Best single-classifier configuration | Best ensemble designs | Best ensemble accuracy | Best accuracy difference |
|---|---|---|---|---|
| ringnorm | 97.66 (SVM) | SVM: simple average, *K*=5 | **95.27** | -2.39 |
|  |  | OBTC: WMR (adaptive) all, *K*=5 | 91.67 |  |
|  |  | w/*k*-NN: DCS-LA (w/priors), *K*=5 | 93.64 |  |
| splice | 92.98 (OBTC) | SVM: DCS-LA (no priors), *K*=5 | 86.49 | +0.26 |
|  |  | OBTC: DCS-LA (no priors), *K*=5 | **93.24** |  |
|  |  | w/*k*-NN: DCS-LA (no priors), *K*=7 | 87.53 |  |
| twonorm | 97.85 (w/*k*-NN) | SVM: simple average, *K*=5 | **96.31** | -1.54 |
|  |  | OBTC: WMR (adaptive), *K*=7 | 90.22 |  |
|  |  | w/*k*-NN: WMR (adaptive) all, *K*=5 | 94.03 |  |
| waveform | 90.10 (SVM) | SVM: WMR (adaptive) all, *K*=5 | 84.39 | -3.20 |
|  |  | OBTC: WMR (adaptive) all, K=7 | 86.43 |  |
|  |  | w/*k*-NN: WMR (adaptive) all, *K*=5 | **86.90** |  |

## 6. Discussion

The results from Table 8 clearly demonstrate the overall superior performance of the "adaptive" WMR model. Both the "adaptive" and the "static" versions of the WMR model show improved performance compared to the simple majority voting, as well as the maximum rule, for all three types of classifier ensembles (SVM, OBTC, w/$k$-NN).

The "adaptive" version of WMR also exhibited better performance compared to the simple averaging rule, as well as the weighted averaging rule using LSE-trained weights, i.e., both soft-output combination models. Thus, the "adaptive" WMR model performs equally well or better than simple soft-output averaging combination rules.

With regard to weighted versus non-weighted combination rules, all three weighted combination rules, i.e., the two WMR and the LSE-trained weighted average, have been clearly proven better than the non-weighted hard-output combination rules (i.e., maximum and simple majority). Moreover, in the overall evaluation, the "static" WMR outperformed the Bayesian-based combination rule using priors ("STD: DCS-LA (with priors)"). This essentially means that the WMR model is a very effective way of exploiting information about the classifiers' competencies, even when this information refers to global (i.e., prior) and not localized (i.e., posterior) probabilities. The overall performance of the WMR improved significantly when local accuracy estimates was used in the "adaptive" version of the model, reaching the top-ranking position over all the other combination rules, including the best soft-output ("STD: LSE-weighted average") and the best Bayesian-based ("STD: DCS-LA (no priors)") combination rules.

The w/Borda rankings from Tables 5 through 8 also demonstrate the robustness and stability of the each combination rule. For small values of standard deviation (close to one) in the corresponding w/Borda mean ranks, the relative ranking position of a combination rule against the others remains more or less the same. Thus, the "static" version of the WMR exhibited a consistently lower ranking position compared to the corresponding "adaptive" WMR model in general. Likewise, the "adaptive" WMR model is more stable than almost all the other combination rules (except maximum), including the LSE-weighted average, which exhibits more or less the same consistency and robustness as the "adaptive" WMR but with lower relative ranking.

In terms of the overall performance of the combination rules, results from Tables 5 through 7 demonstrate that in all cases the best combination rules increased the average success rates (Table 4) of the classifier pool significantly, up to +22% (mainly in the "ringnorm" and "splice" datasets), with larger relative improvements as the size of the ensemble increased from five to seven members, for all the three types of base classifiers. Furthermore, Table 9 shows that the performance of the best ensemble designs closely matched the performance of the corresponding best single-classifier configuration and even surpassed it ("splice" dataset). Although the WMR rule was not always selected as the best ensemble design, its overall performance and the top-ranking positions in Table 8 clearly demonstrate that it is inherently robust and consistently efficient.

The general behavior of almost all the ensembles was consistent with the theoretical background and experimentally verified the assertion that combining even *moderately* independent experts results to the improvement of their individual competencies [1]. Previous studies [61] have shown experimental evidence that optimal combination of SVM classifiers can be achieved through linear combination rules. Ensembles of SVM or other type of robust classifiers, as a combination of multiple simpler models, each using a 1/K portion (subspace) of the feature space of the dataset instead of a single classifier of the same type for the complete feature space, can be used to reduce the overall training effort significantly. In particular, for the SVM model case, kernel evaluation employs inner product between vectors, i.e., its complexity is directly proportional to the dimensionality (number of features) of the input vectors. Thus, feature space reduction, from $F$ to $F/K$ features, results in significant decrease in the overall complexity during training. A similar approach has also been examined in [61], where an ensemble of SVM classifiers has been used, trained with small training sets, instead of a single SVM trained with one large training set. Furthermore, there is evidence that such ensembles of kernel machines are more stable than the equivalent kernel machine itself and that their model need not be more complex than a simple linear combination of its member outputs [61], which is consistent with the theoretical assertion of the WMR formulation as the optimal aggregation model for any $n$-label voting task (see section 2.3). This reduction in training time, of course, has to be compared to the additional overhead of calculating a combination rule for every output vector from the classifier pool, as well as the total training time of the $K$ SVM classifiers. This is one of the main reasons for preferring very simple, linear aggregation schemes with non-trained weights, such as the WMR, for the design of robust classifier ensembles.

It should be noted that using simple linear combination models, including weights that do not require iterative training, can be extremely useful in applications that require parallel and/or on-line updating. In the case of WMR, the combination rule is fully parallelizable, even when using local accuracy estimates in the weighting formula, since they are based on histogram calculations and not on iterative off-line optimization of the weights. Furthermore, the updating of the histogram can also be realized on-line, simply by adding new evaluation results as the model runs on new input data (without any new re-training of classifiers), and only if needed, i.e., when the new data invalidate the statistics of the previous histogram estimations.

## 7. Conclusions

In this paper, a game-theoretic framework for combining classifiers has been proposed. The adapted WMR has been, for the first time, presented as an alternative approach to design simple and efficient ensembles of voting classifiers, even when the conditional independence assumption is only moderately satisfied via feature subspace methods. Experimental comparative results have shown that such simple combination models for combining classifiers can be more efficient than typical rank-based and simple majority schemes, as well as simple soft-output averaging schemes in some cases. Moreover, when the weighting profiles required in the WMR are associated with the posterior (localized), instead of the prior (global), approximations of the classifiers' accuracies, the resulting ensemble can outperform many commonly used combination methods of similar complexity. The use of simple linear combination models that employ analytically computed weights may provide the necessary means to apply multi-expert classification schemes in parallel implementations with on-line updating capabilities. Therefore, the WMR can be asserted as a simple, yet effective tool in the palette for combining classifiers in an optimal, adaptive and theoretically well-defined framework.

## Acknowledgements

The authors wish to thank professor Sergios Theodoridis, Dept. of Informatics & Telecommunications, Univ. of Athens (NKUA/UoA), for his contribution in the early stages of this work.